# MetaHOPE: Metaphor Translation Evaluation Framework Investigating Open-Source LLMs and State-of-the-Art Neural Translation Models

**Jiahui Liang**[1], **Lifeng Han**[2,3]
[1]Centre for Linguistics, Humanities, Leiden University, NL
[2]LIACS, Leiden University, NL
[3]BDS, Leiden University Medical Centre, NL
j.h.l.jiahui@hum.leidenuniv.nl | l.han@lumc.nl

## Abstract

In this paper, we propose MetaHOPE, an error severity-aware annotation framework for evaluating metaphor translations. Metaphors present challenges for machine translation (MT) and natural language understanding and processing (NLU, NLP), because it presents the features of semantic complexity, contextual dependency, and cultural embeddings that can lead to ambiguity issues for NLP models. To investigate how state-of-the-art NLP models perform on translating metaphors, we select three representative systems, i.e., GoogleMT, GPT5.4, and Hunyuan-7b as Neural MT (NMT) models and LLMs. We used two human-annotated metaphor corpora, including VUAMC and PSUCMC, for English-to-Chinese and Chinese-to-English translation purposes. The original corpora we used are monolingual, where we carried out error annotation using the MetaHOPE framework and also produced the human post-edited gold reference for *bilingual use as a new resource*. We believe the MetaHOPE evaluation framework for metaphor translation annotation, the parallel corpora resources, and the error analysis on SOTA automatic translation models can be useful and shed some light for the field of metaphor translation study.[1]

## 1 Introduction

Metaphors are pervasive in everyday discourse and serve as an essential cognitive tool, enabling people to understand and communicate abstract, complex, and unfamiliar concepts through more concrete and familiar experiences. For example, economic indicators may "soar" or "plummet", governments may "fight" inflation, and negotiations may "reach a dead end". These expressions draw on concrete experiences of movement, conflict, and space to convey meanings that extend beyond literal language (Lakoff and Johnson, 1980; Smedinga et al., 2023). Beyond their semantic complexity, metaphors are also culturally embedded, and their interpretation often requires contextual awareness, sociocultural knowledge, and conceptual reasoning. As a result, they pose challenges for both machine translation (MT) and broader natural language understanding and processing (NLU, NLP) tasks.

Recent advances in neural MT (NMT) and large language models (LLMs) have substantially improved translation quality, with some systems achieving performance comparable to human translators on general translation benchmarks (Kocmi et al., 2025). However, such improvements do not necessarily extend to metaphor translation (Han et al., 2026). Karakanta et al. (2025) report metaphor translation accuracy rates of only 64-80%, while Wang et al. (2024) find that around 20% of metaphorical expressions remain non-equivalent in translation. A major *source* of error is overly literal translation, particularly for multi-word expressions (MWEs) such as idioms and collocations, where models often fail to capture the intended figurative meaning (Bhatia et al., 2023, 2024; Han et al., 2024). Therefore, to better understand the gap between general MT performance and metaphor translation performance, it is necessary to *systematically analyze metaphor translation errors*. In addition, existing studies mainly focus on translation strategies (Pedersen, 2017; Zajdel, 2022; Li and Chen, 2025) or translation quality on equivalence, fluency, emotional effect, and authenticity (Wang et al., 2024). However, *fine-grained error analysis* remains limited. Karakanta et al. (2025) classify issues into meaning, form, and omission, but this framework is relatively coarse-grained and does not address severity.

[1]Accepted by ICNLSP2026: 9th International Conference on Natural Language and Speech Processing. resources shared at https://github.com/Jiahui84/MetaHOPE

To address this gap, this study adapts the HOPE framework (Gladkoff and Han, 2022) for metaphor translation evaluation, forming a new framework called **MetaHOPE**. Originally developed as a lightweight version to Multidimensional Quality Metrics (MQM) (Lommel et al., 2014, 2024; Gladkoff et al., 2025), HOPE reduces annotation complexity through a smaller set of error categories and a severity-based scoring scheme. Building on this design, we develop a metaphor-oriented annotation framework that enables the systematic identification and severity assessment of metaphor translation errors. Adapted from HOPE, our project develops a metaphor-oriented annotation framework consisting of five error categories: Impact, Style, Mistranslation, Required Adaptation Missing, and Proofreading Error, together with a five-level severity scale. Using this framework, the study investigates: ***RQ-1)*** What types of metaphor translation errors are produced by different MT systems? **RQ-2)** How do the frequency and severity of errors vary across systems and translation directions (EN-ZH and ZH-EN)?

The empirical results of our pilot study (phase-1) show that our human annotators' agreement levels for [GoogleMT, GPT-5.4, Hunyuan-LLM-7B] are [0.536, 0.726, 0.333] for Pearson's correlation, and [76.9%, 70.8%, 61.5%] for exact agreement. Metaphor translation errors are demonstrated as the main cause of translation errors, occupying [91.7%, 93.8%, 61.8%] error ratios of the three translation systems, respectively. We further qualitatively clustered the MT errors and interesting phenomena into distinct categories.

## 2 Background and Related Work

### 2.1 Metaphors and Translation

Traditional metaphor translations focus on isolated linguistic expressions or rhetorical ornaments. The research areas include the translatability of metaphors, translation procedures, metaphor substitution, and the question of whether the metaphorical image should be preserved (Newmark, 1988; van den Broeck, 1981). Solutions of metaphor translation can be metaphor to same/different metaphor, simile, paraphrase, or deletion (Toury, 2012).

Later, there is the cognitive turn on metaphor understanding and translations, which underlines that metaphor translation shall be conceptual mapping from source to target language, not only on the lexicon level, or stylistic embellishment (Schäffner, 2004; Lakoff and Johnson, 1980). For instance, Hong and Rossi (2021) carried out a survey on cognitive perspectives on metaphor translation, where the authors discussed that the cognitive approach offers insight for cross-cultural communication, using English-Chinese and French-Chinese as distant languages.

Previous studies on metaphor machine translation, including Wang et al. (2024), Dorst (2023) and Karakanta et al. (2025) have largely relied on *sentence-level* translation and evaluation. However, translation scholars have criticized sentence-level MT evaluation, arguing that translations that appear acceptable in isolation may become inappropriate or inaccurate when broader discourse context is considered (Castilho et al., 2020). Hong and Rossi (2021) also emphasized that metaphor translation shall go beyond sentence-by-sentence mapping, and discussed the potential of combining cognitive theory and translation theories. Aligning with this development, in the MetaHOPE design, we carry out the translation at the context-aware document-level first, then extract the translated sentence for annotation, and the annotators are given the context for awareness.

### 2.2 Language/Domain Specific Studies

There are language or domain-specific studies on metaphors and their translations. For instance, from Serbian to English Milenković et al. (2024) study the influence of translation on perceived metaphor features, including Metaphoricity, quality, aptness, and familiarity on both the source and target sides. It covered 55 Serbian metaphors translated into English using the A is B form.

Meanwhile, Khalifah and Zibin (2022) carried out Arabic-English metaphor translation from a cognitive linguistic perspective, using some evidence from Naguib Mahfuz Midaq Alley and its translated version.

Focus on the science domain, Shuttleworth (2017) examines how figurative language in popular science articles functions across languages and bridges the gap between metaphor research and translation studies, specifically in neurobiology and biotechnology. This work challenges the notion that scientific language is purely literal, arguing that metaphor is a vital component of transferring complex scientific concepts to the public. They develop new, theoretically nuanced procedures to describe how translators navigate and adapt metaphorical language across different linguistic and cultural contexts. Similarly, the work by Smedinga et al. (2023) studies metaphors as tools for understanding in science communication among experts and to the public. In our work, for MetaHOPE, we use English-Chinese bidirectional studies and focused on the news domain for proof-of-concept.

### 2.3 LLMs on Metaphor Translation

Using LLMs for metaphor translation is still an under-explored area. Wong and Xu (2025) conducted a bibliometric analysis of metaphor research in Translation and Interpreting Studies (TIS) based on 1,023 publications from 1964 to 2023. They found that machine translation represents only 0.68% of translation-related metaphor studies, highlighting a substantial research gap in this area.

Recent work has begun to explore the use of large language models for metaphor-related tasks. For example, Şen Bartan et al. (2026) employed GPT-4o with few-shot prompting to detect conceptual legal metaphors in English–Turkish HUDOC judgments and analyze conceptual shifts across translations. However, their focus was metaphor identification and conceptual labeling rather than evaluating the quality of metaphor translation outputs. In contrast, the present study investigates how different MT/LLM systems render metaphors in translation and evaluates translation quality using a metaphor-adapted MetaHOPE framework

In parallel, some NLP work has begun to address **figurative language translation** using large language models. Donthi et al. (2025), for example, investigated idiomatic translation and proposed semantic idiom alignment methods to improve LLM handling of non-literal expressions, finding that semantic-level alignment better preserved figurative meaning and cultural authenticity than direct prompting approaches. However, their focus was idiomatic translation generation rather than systematic evaluation of metaphor translation quality.

The most relevant work to ours is from Li and Chen (2025). This work examined the translation of metaphor-related words (MRWs) by human translators, NMT, and LLMs, combining translation product analysis with think-aloud protocols and quality assessment. Their findings suggest that LLMs produce translation strategies more similar to human translators than conventional NMT systems, although performance remains inconsistent for novel metaphors. However, their analysis focuses primarily on translation strategies and MRW-level behavior rather than systematic evaluation of metaphor translation quality. In contrast, the present study investigates metaphor translation outputs at the segment-level (with context) using a metaphor-adapted MetaHOPE framework (a dedicated error taxonomy) to analyze fine-grained error distributions across MT and LLM systems.

## 3 MetaHOPE Methodology

**Resources.** The annotation guidelines, processed data, and related resources are publicly available at `https://github.com/Jiahui84/MetaHOPE`.

As shown in Figure 1, from left-to-right and top-to-down, the overview of MetaHOPE framework including the following steps:

- 1) Text formatting and preprocessing from VUAMC and PSUCMC corpora. This step includes a) plain text extraction, and b) CSV file preparation, including word id, if it is a metaphor, POS, token-position, etc. We present an example table in Figure 6 and 7 for the Chinese and English corpora, formatted accordingly.

- 2.1) Machine translation (MT) on the two source texts to the target languages ona full document-level for context-awareness, English-to-Chinese and Chinese-to-English using three selected systems: GoogleMT, GPT5.4, and Hunyuan-llm-8b as representatives of state-of-the-art NMT systems, LLMs, and state-of-the-art performing system at the annual WMT shared task on MT (Kocmi et al., 2025).

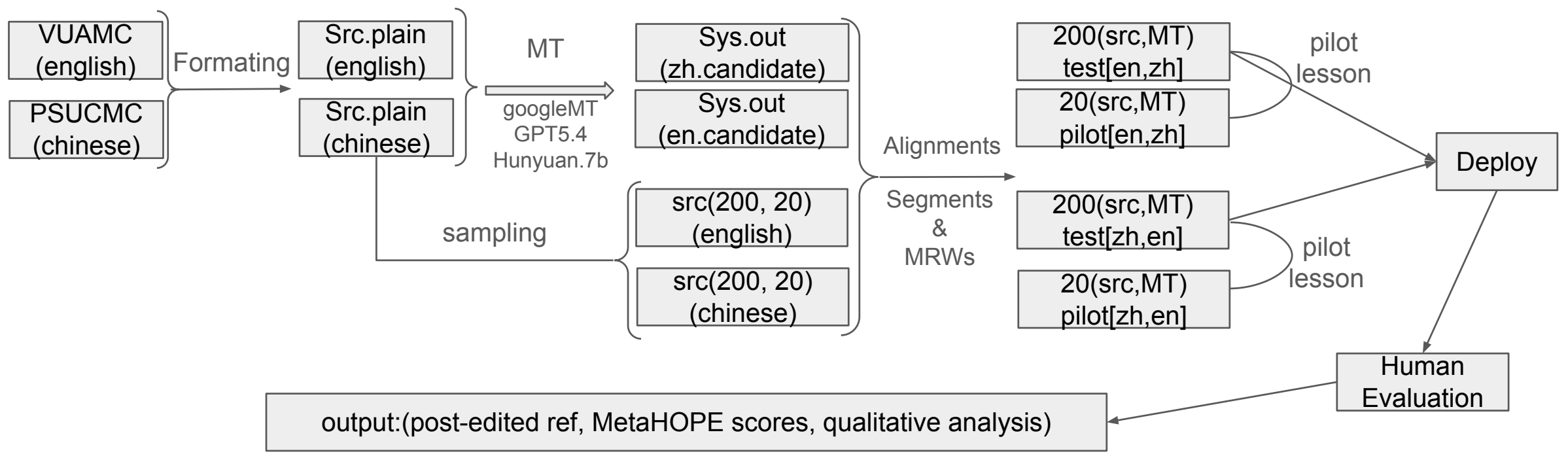

Figure 1: MetaHOPE Framework: Metaphor corpus preparation, MT, Aligning segments and metaphor-related words (MRWs), and post-editing with annotations on pilot (development) and test sets.

- 2.2) In parallel with 2.1, we sample 20 and 200 segments from each of the two corpora as a pilot study and system testing set, respectively.

- 3) a) Manual alignment of the four data sets 2 x (20, 200) segments to the system translation outputs to find parallel translations for English-Chinese and Chinese-English pairs. b) Manual alignment of metaphor-related words (MRWs) in the target MT outputs towards the source-side metaphor words (more details in Section E.3).

- 4) Pilot studies are carried out on the two sets of 20 segments from the two translation directions. These lessons are used to discuss the metaphor translation error annotation guidelines, resolving annotators' disagreement, refine the annotation policies, for the next stage role-out.

- 5) Large-size human annotation on translation outputs from three MT systems on metaphor related errors producing: a) post-edited human reference for each translation direction, b) MetaHOPE score table generation on three systems, and c) qualitative analysis on error types and MT behaviors on metaphor translation.

Regarding error categories in MetaHOPE, we limit into the following 5 types instead of the original 8 used by HOPE metric (updated definition in Section J.2 for our **Phase-2** experiments):

- Impact (IMP): Over-literal translation; structural shifts affecting emphasis.
- Required Adaptation is Missing (RAM): Missing cultural or idiomatic adaptation of metaphor.
- Mistranslation (MIS): Meaning mismatch; incorrect interpretation of metaphor.
- Style (STL): Loss of metaphorical effect, imagery, or emotional tone.
- Proof-reading error (PRF): Awkward or unnatural expression (not reflecting meaning).

The principle design of this five categories is according to the existing metaphor-focused studies and their mapping to the original HOPE categories. The HOPE framework defines eight error categories. However, not all of them are equally relevant for metaphor translation. Based on existing literature on metaphor translation, common error types include overly literal translation, meaning mismatch, loss of metaphorical (rhetorical and aesthetic) effect, emotional shift, structural changes (e.g. active–passive alternation depending on discourse context), omission (when meaning is lost), and unauthentic expression. These error types are mapped onto the HOPE framework to adapt it for metaphor translation. We further list examples of each of these five error categories from MetaHOPE in Table 8.

Categories such as TRM (Terminology), PRN (Proper Name), and UGR (Ungrammatical) are excluded, as they operate at a different level from metaphor processing. TRM mainly concerns terminology consistency, PRN relates to the correct translation of named entities, and UGR captures grammatical well-formedness. While these issues may affect overall translation quality, they do not directly explain how figurative meaning is interpreted, adapted, or expressed. In addition, these categories are not explicitly reflected in commonly discussed error types in the metaphor translation literature.

For quantitative scoring on error penalties for each error type, we keep the 5 severity levels, i.e. minor, medium, major, severe, and critical, but use the following score alignment (2, 4, 6, 8, 10) instead of the exponential score range used by original HOPE ($2^x$, 1, 2, 4, 8, 16) (Gladkoff and Han, 2022). The rationale is that we think the original score range can be very sparse, e.g., the same error can be labeled by different annotators as 4 (medium) or 16 (critical) which potentially leads to higher disagreement.

In addition, we describe *how many percent of errors are from metaphors and how many from non but in the sentence, in our data.*

## 4 Experimental Evaluation

### 4.1 Data Preprocessing

We list the details on how we extracted the sampled data according to part-of-speech (POS), and the segmentation steps, Section E including data extraction and filtering. The metaphor-related word alignment from the source to the target language is carried out using the guidelines in Section E.3.

### 4.2 Pilot Study on Development Set

Two Annotators' Backgrounds: Annotator-A is a PhD candidate in linguistics and translation studies, who holds an MA in digital humanities and a BA in translation. Annotator-B is a Master's student majoring in translation and interpreting studies (MTI). The Annotations are carried out independently with the instruction manual.

The Pilot-Study Mile-stone-II is described in Section J with six annotators.

#### 4.2.1 Annotator Agreements

Inter-annotator agreement was evaluated using Krippendorff' s $\alpha$ and quadratic weighted Cohen's $\kappa$, treating MetaHOPE severity scores as ordered penalty levels (0, 2, 4, 6, 8, 10). Segment-level (SEGS) agreement was computed by summing IMP, RAM, MIS, STL, and PRF into SEGS for each metaphor instance as in Table 1. More detailed per-error-type agreement is listed in Table 11. From these two tables, the strongest agreement appears for MIS, especially for Hunyuan-LLM-7B. However, overall $\alpha/\kappa$ values are low because Annotator B applied much higher total penalties than Annotator A. Exact agreement is high because most cells are zero, but $\alpha/\kappa$ reveal the severity-bias problem more clearly.

Although exact agreement was relatively high, $\alpha$ and $\kappa$ were modest, suggesting systematic severity differences between annotators, particularly with Annotator B assigning substantially higher penalties.

GPT-5.4 has the highest annotator consistency (r = 0.726), GoogleMT moderate agreement, Hunyuan lower agreement, which is probably because it generated more varied metaphor outputs, making annotation harder or more subjective. This is an interesting finding from the pilot study: harder-to-interpret translations may reduce annotator consistency.

#### 4.2.2 MetaHOPE Error Statistics

There are 65 lines of translation annotation on 65 metaphors with 20 unique sentences for English-to-Chinese and 32 for Chinese-to-English, some sentences having multiple metaphor words. From Annotator-A, the summary of full sentence-level error penalty (**EPS**), which includes metaphor level/component penalty, is displayed in Table 2. Correspondingly, the summary of error penalty on metaphor (**EPM**) only across each error type on the 65 lines is shown in Table 3 for the three tested systems, where the Rotio (M/S) is the Ratio of EPM/EPS, i.e., the value of error scores of metaphors divided by the value of error scores of the full sentences. From this Ratio(M/S), we can see that the metaphor caused errors occupy around 91.7% to 93.8% for GoogleMT and GPT-5.4, i.e., the major cause of automated translation er-rors. While Hunyuan-LLM-7B has metaphor-caused errors 61.8% as a main source (>50%), it also has many other error types, including hallucination, which we will discuss more in the next section on qualitative analysis/categorization (Section 4.2.3).

Table 1: Segment-level inter-annotator agreement on summed MetaHOPE penalties (SEGS).

| System | $\alpha$ | Weighted $\kappa$ | Pearson $r$ | Exact Agree. | A/B Totals |
|---|---|---|---|---|---|
| GoogleMT | 0.235 | 0.305 | 0.536 | 76.9% | 22 / 124 |
| GPT-5.4 | 0.304 | 0.348 | 0.726 | 70.8% | 30 / 150 |
| Hunyuan-LLM-7B | 0.105 | 0.191 | 0.333 | 61.5% | 42 / 226 |

#### 4.2.3 Qualitative Categorization of Errors

We categorize some of the error phenomena from the three systems, which can be useful for further research on this topic.

1) *GoogleMT and GPT5.4 are more fact-tracking, while Hunyuan-llm-7b is more flexible in generated translation.* This flexibility sometimes generates more native translation, while other times it can be hallucinatory, e.g., by adding extra information or losing some source meaning (addition or reduction). Example-1:

- Src: An organisation that doesn' t change fossilises. (**fossilises**; V)
- GoogleMT: 一个不**改变**的组织就会**僵化**。(strict word-order translation from the source)
- GPT5.4: 一个不**变化**的组织会**僵化**。(same like GoogleMT)
- Hunyuan-7b: **变革**是必要的，否则组织就会**僵化停滞** (change is necessary; otherwise, the organisation will fossilise.)

In Example-1, Hunyuan-7b used a more flexible order than the original English word order, which actually makes it sound more native. The other two translations are more like literal translations, keeping strict word tracking in translation (rigid). However, Hunyuan-7b's output is also debatable, since "变革是必要的 (Change is necessary)" is not exactly included in the source sentence (必要的—necessary). Example-2:

- Src: She **pledged** that the Government would safeguard those that did not opt for *trust status*, but she expected this to be a minority. (pledged; V)
- GoogleMT: 她**承诺**政府将保障那些未选择**信托地位**的诊所的权益，但她预计这只是少数。(literal translation of "trust status")
- GPT5.4: 她**承诺**，政府将保障那些不选择**信托地位**的医院，但她预计这将只是少数。(literal translation of "trust status")
- Hunyuan-7b: 她**相信**... 政府也会保护那些未选择**自主管理**的医院，但这类医院只会占少数。(adapted translation of "trust status" to "self-governing"; but "pledged" not translated properly)

In this example, on the one hand, "信托地位" from GoogleMT and GPT5.4 is a literal translation of "trust status" while "自主管理" from Hunyuan-7b is more down-to-earth (native-like) Chinese that most people can understand better. Even though "信托地位" also exists in Chinese, it is more like a borrowing word, and lay people will not understand what it means. So Hunyuan-7b did a better job on this foreign concept translation/localization. On the other hand, "相信" is not accurate enough from Hunyuan-7b. Interestingly, current ChatGPT (using GPT5.5) can explain this literal translation very well from GPT5.4 (API), when we gave the following prompt:

> «I am doing a project on metaphor translation investigation using three different models - GoogleMT/GPT5.4/Hunyuan-llm. I need some feedback when I am doing annotation for human gold standard reference preparation from English to Chinese. From the following sentence, "She pledged that the Government would safeguard those that did not opt for trust status, but she expected this to be a minority." What does "trust status" mean?»
> The response of ChatGPT is: «"她承诺政府会保障那些不选择转为信托制的学校，但她预计这只会是少数。" for school, or "成为信托机构的资格/地位" and "转为 NHS 信托机构" for healthcare.»

Example-3:

- Src: Earnings were level at 17.5p a share. (level; AJ)

Table 2: MetaHOPE Sentence-level Penalty **EPS** Distribution Across Three MT Systems (annotator-A)

| System | IMP | RAM | MIS | STL | PRF | SEGS | Per Sentence |
|---|---|---|---|---|---|---|---|
| GoogleMT | 0 | 0 | 20 | 0 | 4 | 24 | 1.2 |
| GPT-5.4 | 2 | 0 | 18 | 0 | 12 | 32 | 1.6 |
| Hunyuan-LLM-7B | 4 | 0 | 64 | 0 | 0 | 68 | 3.4 |

Table 3: MetaHOPE Metaphor-Level Penalty Score **EPM** Distribution on Three MT Systems(annotator-A)

| System | IMP | RAM | MIS | STL | PRF | SEGS | Per Sentence | Per Metaphor | Ratio (M/S) |
|---|---|---|---|---|---|---|---|---|---|
| GoogleMT | 0 | 0 | 16 | 0 | 6 | 22 | 1.1 | 0.338 | 0.917 |
| GPT-5.4 | 2 | 4 | 16 | 0 | 8 | 30 | 1.5 | 0.462 | 0.938 |
| Hunyuan-LLM-7B | 4 | 0 | 36 | 0 | 2 | 42 | 2.1 | 0.646 | 0.618 |

- GoogleMT: 每股收益**持平于** 17.5 便士。(correct translation)
- GPT5.4: 每股收益**持平，为** 17.5 便士。(correct translation)
- Hunyuan: 每股收益**为** 17.5 便士 (meaning-loss translation)

In this Example-3, Hunyuan-7b *lost* the "level" information only saying "收益为", which is an important feature in finance context, i.e., not more or less, but the same — unchanged compared to the previous reporting period. Meanwhile, the other two systems used "收益持平" which indicates the "level" is the same as the previous report.
2) *Term Translation Inconsistency (across systems).* There are situations of inconsistency in term translation from the same translation system. For instance, GoogleMT sometimes translates "FT-SE" into "富时 100" while other times just keeps "FT-SE" as it is, with two examples below. Example-1:

- Src: Prices have remained high-indeed the FT-SE index has risen another 55 points since then —allowing even the most passive of private investors, including unit trust holders, to take advantage of the market.
- GoogleMT: 价格一直保持高位——事实上，自那以后，*FT-SE* 指数又上涨了 55 点——这使得包括单位信托基金持有者在内的最被动的私人投资者也能从市场中获利。(kept "FT-SE")

Example-2:

- Src: The change in employment may not always be so favourable as yesterday's either, but the market is now starting to feel bullish and looking for a FT-SE of 3,000 by next year.
- GoogleMT: 就业形势的变化或许不会总是像昨日那样乐观，但市场目前开始看涨，并预期富时 *100* 指数明年将达到 3000 点。(used 富时 *100*)

3) *Metaphorical word-alignment* — metaphor-word to none (general issues). There are situations when there is no clear target word mapping to a source metaphor word. In Example-2 of the last Inconsistency issue, GoogleMT used "看涨" for "feel bullish", though there is no exact/separate translation of "feel". However, one of the individual entry of this source English sentence focuses on (feel; V). This reflects the issue of "word-level" metaphor translation.

Table 4: Five-annotator metaphor-level agreement by MetaHOPE error type, pooled across GoogleMT, GPT-5.4, and Hunyuan-7B.

| Type | Ordinal $\alpha$ | Mean weighted $\kappa$ | Exact agreement |
|---|---|---|---|
| IMP | 0.038 | 0.063 | 65.6% |
| RAM | 0.116 | **0.114** | 76.0% |
| MIS | **0.141** | 0.084 | 78.1% |
| STL | −0.035 | −0.027 | 82.3% |
| PRF | 0.107 | 0.050 | **86.5%** |

4) *The selection of Chinese interchangeable words.* There are often inter-changeable words in modern Chinese, however, there are indeed new regulations to guide their use, as reflected by the Act "Legislative Drafting Technical Standards (Trial Implementation) (II) (Document No. [2011] 5 issued by the Legislative Affairs Commission of the Standing Committee of the National People' s Congress)|《立法技术规范 (试行)(二)》(全国人大常委会法工委发 [2011]5 号)。" [2] For instance, the words "做出" and "作出" in the following example:

- Src: She recalled a promise made by Mr Major when he became Prime Minister: that he would work for a nation at ease with itself. (ease; N)
- GoogleMT: 她回顾了梅杰先生就任首相时**做出**的承诺：他将致力于建设一个安心自在的国家。
- GPT5.4: 她回忆起梅杰先生出任首相时**作出**的承诺：他将致力于建设一个与自身和谐相处的国家。
- Hunyuan-7b: 她回忆起约翰·梅杰在担任首相时曾承诺要打造一个**让民众安心的医疗体系**
- Ref: 她回顾了梅杰先生就任首相时作出的承诺: 他将致力于建设一个安心自在的国家。

## 4.3 Pilot Study Revisit (Phase-2)

Following our first pilot study, we revised annotation guidelines using our findings and recruited more annotators for the 2nd round annotation experiments on the same Dev dataset, which work flow is shown in Figure 9. In the second round, we recruited overall

[2] Post from Chinese University of Petroleum, Beijing, accessed 2026.8th June.https://www.cup.edu.cn/yww/jpbl/d0f2ffb8c08a4f3cb74782ed39a50fef.htm

Table 5: Three-annotator sentence-level agreement on the total HOPE penalty across the five error categories.

| **System** | **Ordinal $\alpha$** | **Mean weighted $\kappa$** | **ICC(2,1)** | **ICC(2,$k$)** | **Exact score agreement** | **Exact error-presence agreement** |
|---|---|---|---|---|---|---|
| GoogleMT | **0.232** | 0.146 | **0.191** | **0.415** | 45.0% | 50.0% |
| GPT-5.4 | 0.111 | 0.049 | 0.067 | 0.178 | **55.0%** | 55.0% |
| Hunyuan-7B | **0.278** | **0.182** | 0.161 | 0.365 | 30.0% | **60.0%** |

Table 6: Five-annotator metaphor-level inter-annotator agreement for MetaHOPE across the three MT/LLM systems. Agreement is computed over 32 metaphor instances using ordinal Krippendorff's $\alpha$, mean pairwise quadratic weighted Cohen's $\kappa$, exact total-penalty agreement, and exact error-presence agreement.

| **System** | **Ordinal $\alpha$** | **Mean weighted $\kappa$** | **Exact score agreement** | **Exact error presence agreement** |
|---|---|---|---|---|
| GoogleMT | 0.183 | 0.142 | 34.4% | 37.5% |
| GPT-5.4 | **0.221** | **0.241** | **46.9%** | **46.9%** |
| Hunyuan-7B | 0.159 | 0.133 | 9.4% | 15.6% |

Table 7: Three-annotator sentence-level agreement by HOPE error type, pooled across GoogleMT, GPT-5.4, and Hunyuan-7B.

| **Type** | **Ordinal $\alpha$** | **Mean weighted $\kappa$** | **Exact agreement** |
|---|---|---|---|
| IMP | −0.029 | 0.000 | 90.0% |
| RAM | 0.170 | 0.057 | 83.3% |
| MIS | **0.366** | **0.211** | 65.0% |
| STL | −0.035 | −0.031 | 88.3% |
| PRF | −0.041 | 0.000 | 86.7% |

5 annotators for both English<=>Chinese directions. Their backgrounds are: four students from the major of Translation and Interpretation Studies for their master's degrees; one PhD candidate in Linguistics and Translation Studies. Out of the five annotators, two of them only annotated at the metaphor-token level (32 rows), while the other 3 did both on metaphor-token and sentence-level scores (52 rows).

**Metaphor Instance/Token Level.** From the metaphor-level IAA output (Table 6), we can see that 1) GPT-5.4 has the highest metaphor-level agreement. Its outputs appear to give annotators more consistent evidence for deciding whether an error exists and how severe it is. 2) Hunyuan-7B has by far the lowest exact agreement. This supports our earlier qualitative observation (Section 4.2.3): Hunyuan frequently produces freer reformulations. These may sound fluent while also introducing omissions, additions, shifts in agency, or unsupported interpretations, making categorisation more subjective. The $\alpha$ and $\kappa$ values remain low overall. This means that the five annotators do not yet apply the *severity* scheme consistently enough.

The metaphor-level IAA per error type is displayed in Table 4. From this table, we can see that 1) PRF and STL have high exact agreement largely because annotators usually assigned zero. Their low or negative chance-corrected agreement indicates that, when such errors did occur, the annotators did not identify or score them consistently. 2) The most meaningful agreement appears in a) MIS: highest ordinal $\alpha$, suggesting some shared ordering of mistranslation severity; and b) RAM: highest mean weighted $\kappa$, suggesting relatively consistent judgements when cultural or idiomatic adaptation is considered necessary. 3) The weakest operational category is STL. The negative $\alpha$ and $\kappa$ indicate that the occasional non-zero STL decisions are highly inconsistent across annotators.

**Sentence Level.** We first present the three-annotator sentence-level agreement based on the total HOPE penalty across the five error categories in Table 5. In the table: ICC(2,1) estimates reliability when using one individual annotator. ICC(2,k) estimates reliability when averaging the three annotators. The ICC results show that an average of three annotators is notably more reliable than relying on one annotator, particularly for GoogleMT and Hunyuan. We further list the three-annotator sentence-level agreement by HOPE error type, pooled across GoogleMT, GPT-5.4, and Hunyuan-7B, Table 7. At sentence level, **MIS** is clearly the most reproducible category. This suggests that annotators find it easier to *agree on whether an entire translation contains a meaning error* than to agree on subtler impact, style, or naturalness issues. Again, the high exact agreement for IMP, STL, and PRF mostly reflects agreement that no such error was present.

## 4.4 Visualisation of Outcomes

Figures 2–5 provide complementary visualisations of the pilot results, highlighting the relative contribution of metaphor-related errors, the overall penalty burden, the distribution of MetaHOPE error types, and the resulting system-level profiles.

Figure 2 shows the proportion of the total sentence-level penalty that can be attributed specifically to metaphor-related errors. Metaphor-related penalties account for 91.7% and 93.8% of the total penalties for GoogleMT and GPT-5.4, respectively, indicating that metaphor translation constitutes the dominant source of error in these outputs. For Hunyuan-LLM-7B, the corresponding proportion is lower at 61.8%; nevertheless, metaphor-related errors still account for more than half of its total penalty. The lower proportion for Hunyuan should be interpreted together with its larger overall sentence-level penalty, which indicates the presence of substantial non-metaphor-related translation errors as well.

Figure 3 compares the average sentence-level penalty with the average metaphor-level penalty for each system. Hunyuan-LLM-7B receives the highest penalty on both measures, with an average sentence-level penalty of 3.4 and an average metaphor-level penalty of 0.646. GPT-5.4 occupies an intermediate position (1.6 and 0.462), while GoogleMT receives the lowest average penalties (1.2 and 0.338). These results suggest clear differences in the error burden of the three systems within the present pilot sample.

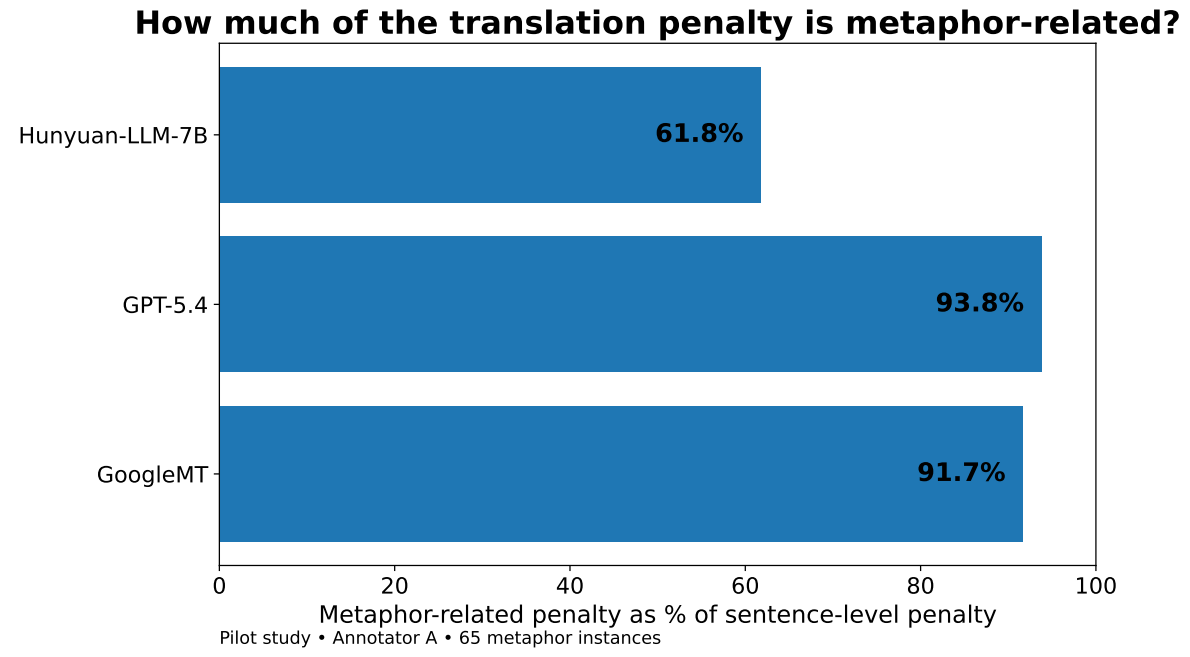

Figure 2: Metaphor Error Ratio.

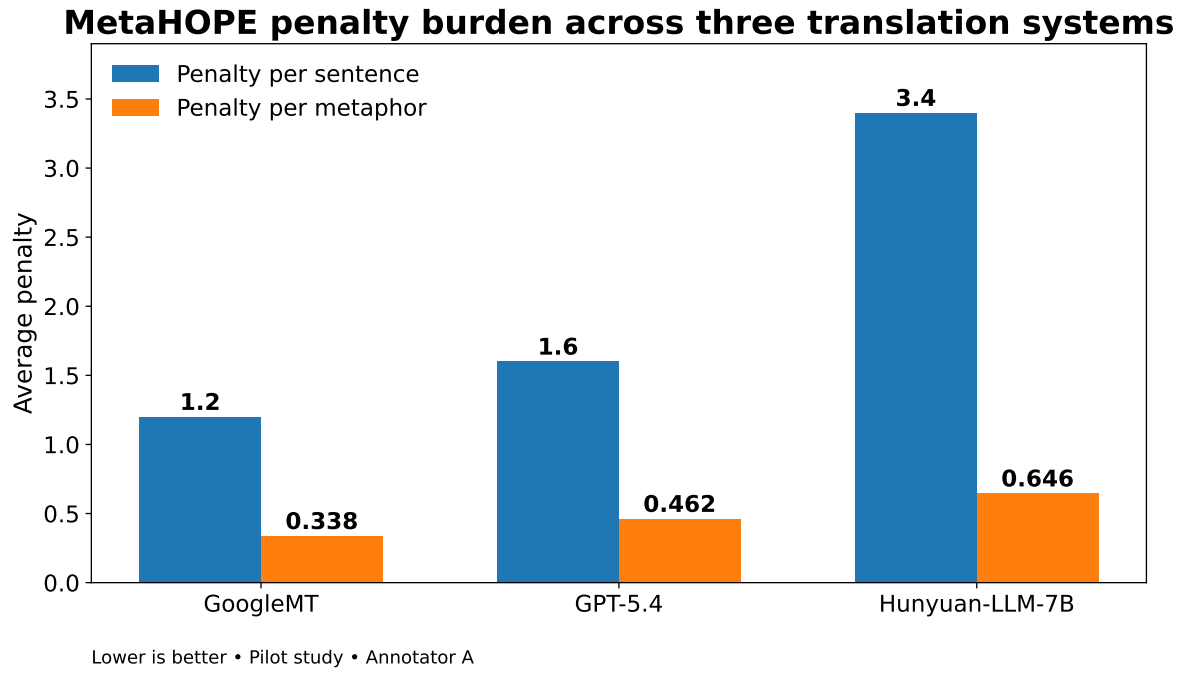

Figure 3: Penalty Comparison.

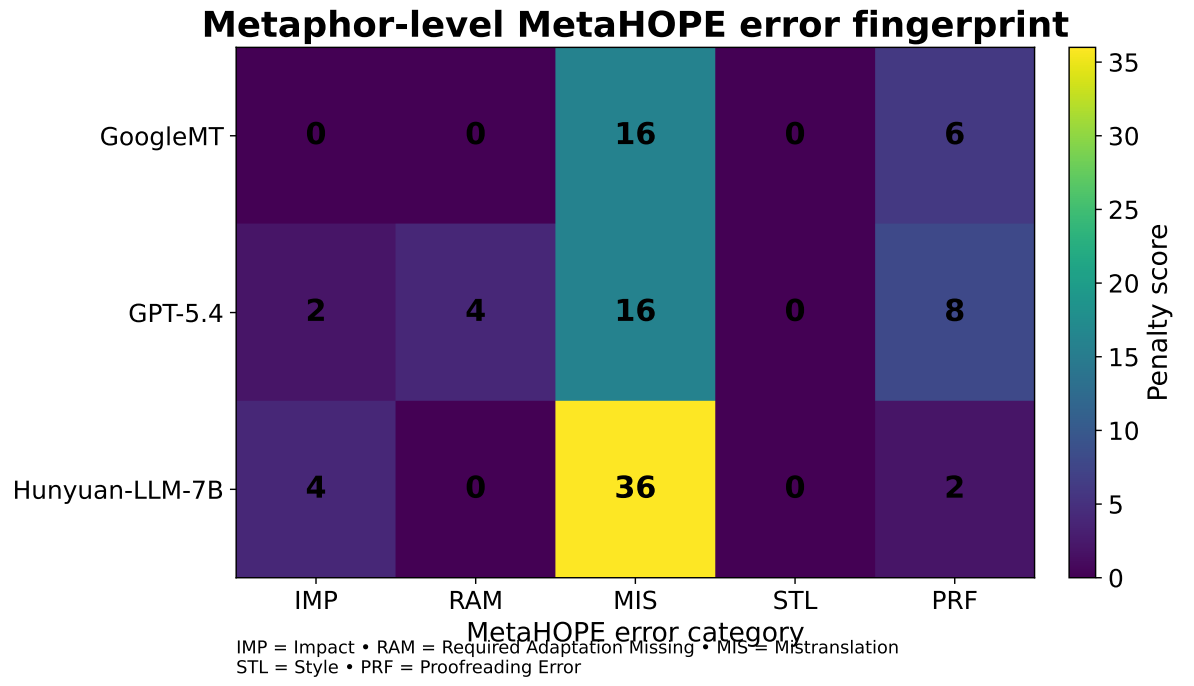

Figure 4: Error Type Heatmap.

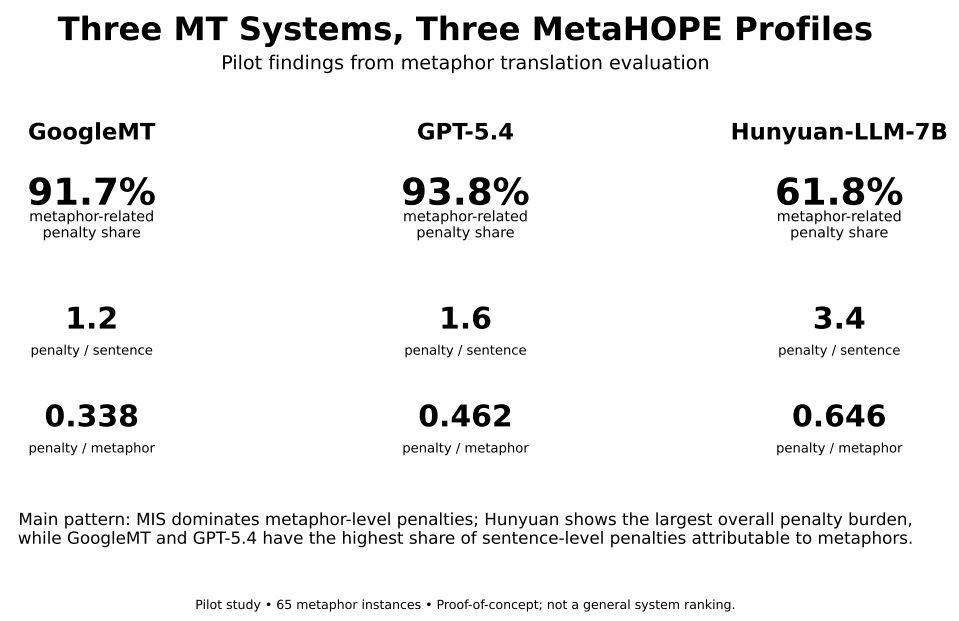

Figure 5: Three System Summary.

Figure 4 further decomposes the metaphor-level penalties according to the five MetaHOPE error categories. Mistranslation (MIS) is the dominant category across all three systems, contributing penalty scores of 16 for both GoogleMT and GPT-5.4 and 36 for Hunyuan-LLM-7B. Other error categories occur less frequently and show more system-specific patterns: GPT-5.4 receives penalties for Required Adaptation Missing (RAM) and Proofreading Error (PRF), whereas Hunyuan-LLM-7B is particularly characterised by its high MIS penalty. No Style (STL) penalties were assigned in this annotation sample.

Finally, Figure 5 brings these measures together to illustrate three distinct system profiles. GoogleMT exhibits the lowest overall penalty burden but a high proportion of metaphor-related error; GPT-5.4 similarly shows a high metaphor-related penalty share, with a moderately higher overall penalty; and Hunyuan-LLM-7B exhibits the highest overall and metaphor-level penalties, while metaphor-related errors constitute a smaller share of its total errors. Taken together, these visualisations suggest that MetaHOPE can reveal not only how much translation error is associated with metaphors, but also how the nature and severity of these errors differ across translation systems. As these results are based on the pilot sample, they should be interpreted as proof-of-concept patterns rather than as a general ranking of the three systems.

## 5 Conclusions

We introduced MetaHOPE, a metaphor-oriented adaptation of the HOPE translation evaluation framework for investigating metaphor translation performance in neural machine translation (NMT) and large language models (LLMs). Motivated by the semantic complexity, contextual dependency, and cultural embeddedness of metaphor, MetaHOPE operationalises metaphor translation quality through five error categories—Impact (IMP), Required Adaptation Missing (RAM), Mistranslation (MIS), Style (STL), and Proofreading Error (PRF)—combined with a severity-aware scoring scheme. In doing so, the framework transforms *cognitively motivated* concerns in metaphor translation, such as conceptual meaning transfer, metaphor remapping, and pragmatic-cultural adaptation, into measurable annotation categories suitable for empirical MT evaluation.

For proof-of-concept, using metaphor-containing news data sampled from the VUAMC and PSUCMC corpora, we conducted a pilot investigation of three representative translation systems: GoogleMT, GPT-5.4, and Hunyuan-LLM-7B, across English–Chinese and Chinese–English translation settings. Preliminary findings suggest that metaphor translation remains a major challenge for current systems. Metaphor-related errors account for a substantial proportion of overall translation penalties, indicating that figurative language remains a key bottleneck despite recent improvements in general MT quality. We also observed systematic differences across systems: GoogleMT and GPT-5.4 tended to preserve source-side wording more conservatively, while Hunyuan-LLM-7B demonstrated greater flexibility and localization ability, although sometimes at the cost of factual consistency or hallucination.

## Limitations

The pilot study revealed that harder-to-interpret metaphor translations may reduce inter-annotator consistency, highlighting the importance of clear annotation guidelines and cognitively informed evaluation criteria.

## Acknowledgments

We thank the anonymous reviewers for valuable comments on our work and the annotators for the human evaluations. This current updated version includes the annotators from pilot-v1: Ziwei Zhang; pilot-v2: Xuemei Deng, Ting Duan, Yang Huang, and Aitian Zhang (alphabetical order).

We thank Prof. Zhengmao Hu for helping us advertise this project for recruiting annotators.

We thank the Chinese Scholarship Council fund for supporting this work.

## References

Abdullah Alabdullah, Lifeng Han, and Chenghua Lin. 2026. Ara-hope: Human-centric post-editing evaluation for dialectal arabic to modern standard arabic translation. In *Proceedings of the 13th Workshop on NLP for Similar Languages, Varieties and Dialects*, pages 157–171.

Mona Baker. 1992. In other words: a coursebook on translation. Technical report, Routledge.

Susan Bassnett. 2013. *Translation studies*. routledge.

Archna Bhatia, Gosse Bouma, A Seza Doğruöz, Kilian Evang, Marcos Garcia, Voula Giouli, Lifeng Han, Joakim Nivre, and Alexandre Rademaker. 2024. Proceedings of the Joint Workshop on Multiword Expressions and Universal Dependencies (MWE-UD)@ LREC-COLING 2024. In *Proceedings of the Joint Workshop on Multiword Expressions and Universal Dependencies (MWE-UD)@ LREC-COLING 2024*.

Archna Bhatia, Kilian Evang, Marcos Garcia, Voula Giouli, Lifeng Han, and Shiva Taslimipoor, editors. 2023. *Proceedings of the 19th Workshop on Multiword Expressions (MWE 2023)*. Association for Computational Linguistics, Dubrovnik, Croatia.

Esperança Bielsa and Susan Bassnett. 2008. *Translation in global news*. Routledge.

Julia Birke and Anoop Sarkar. 2006. A clustering approach for nearly unsupervised recognition of nonliteral language. In *11th Conference of the European chapter of the association for computational linguistics*, pages 329–336.

Julia Birke and Anoop Sarkar. 2007. Active learning for the identification of nonliteral language. In *Proceedings of the Workshop on Computational Approaches to Figurative Language*, pages 21–28.

Sheila Castilho, Maja Popović, and Andy Way. 2020. On context span needed for machine translation evaluation. In *Proceedings of the Twelfth Language Resources and Evaluation Conference*, pages 3735–3742.

John Cunnison Catford. 1965. *A linguistic theory of translation*, volume 31. Oxford university press London.

Andrew Chesterman. 1997. Ethics of translation. *Benjamins translation library*, 20:147–160.

Sundesh Donthi, Maximilian Spencer, Om B. Patel, Joon Young Doh, Eid Rodan, Kevin Zhu, and Sean O'Brien. 2025. Improving LLM abilities in idiomatic translation. In *Proceedings of the First Workshop on Language Models for Low-Resource Languages*, pages 175–181, Abu Dhabi, United Arab Emirates. Association for Computational Linguistics.

Aletta G Dorst. 2023. Metaphor in literary machine translation: style, creativity and literariness. In *Computer-assisted literary translation*, pages 173–186. Routledge.

Serge Gladkoff and Lifeng Han. 2022. Hope: A task-oriented and human-centric evaluation framework using professional post-editing towards more effective mt evaluation. In *Proceedings of the Thirteenth Language Resources and Evaluation Conference*, pages 13–21.

Serge Gladkoff, Lifeng Han, and Katerina Gasova. 2025. Non-linear scoring model for translation quality evaluation. *arXiv preprint arXiv:2511.13467*.

Serge Gladkoff, Irina Sorokina, Lifeng Han, and Alexandra Alekseeva. 2022. Measuring uncertainty in translation quality evaluation (tqe). In *Proceedings of the Thirteenth Language Resources and Evaluation Conference*, pages 1454–1461.

Serge Gladkoff, Angelika Vaasa, Sue Ellen Wright, Ingemar Strandvik, and Lifeng Han. 2026. Looking under the wrong lamppost: On the limitations of automated translation quality estimation. In *Proceedings of the 9th International Conference on Natural Language and Speech Processing (ICNLSP 2026)*, Trento, Italy. To appear.

Pragglejaz Group. 2007. Mip: A method for identifying metaphorically used words in discourse. *Metaphor and symbol*, 22(1):1–39.

Lifeng Han, Kilian Evang, Archna Bhatia, Gosse Bouma, A Seza Doğruöz, Marcos Garcia, Voula Giouli, Joakim Nivre, and Alexandre Rademacher. 2024. Overview of MWE history, challenges, and horizons: standing at the 20th anniversary of the MWE workshop series via MWE-UD2024. *arXiv preprint arXiv:2412.18868*.

Lifeng Han, Najet Hadj Mohamed, Malak Rassem, Gareth JF Jones, Alan F Smeaton, and Goran Nenadic. 2026. Towards a resource for multilingual lexicons: an mt assisted and human-in-the-loop multilingual parallel corpus with multi-word expression annotation. *Language Resources and Evaluation*, 60(2):33.

Wenjie Hong and Caroline Rossi. 2021. The cognitive turn in metaphor translation studies: A critical overview. *Journal of Translation Studies*, 5(2):83–115.

Alina Karakanta, Mayra Nas, and Aletta G Dorst. 2025. Metaphors in literary machine translation: Close but no cigar? In *Proceedings of Machine Translation Summit XX: Volume 1*, pages 276–286.

Lama Khalifah and Aseel Zibin. 2022. Arabic-english metaphor translation from a cognitive linguistic perspective: evidence from naguib mahfuz midaq alley and its translated version. *Babel*, 68(6):860–889.

Tom Kocmi, Ekaterina Artemova, Eleftherios Avramidis, Rachel Bawden, Ondřej Bojar, Konstantin Dranch, Anton Dvorkovich, Sergey Dukanov, Mark Fishel, Markus Freitag, and 1 others. 2025. Findings of the wmt25 general machine translation shared task: Time to stop evaluating on easy test sets. In *Proceedings of the Tenth Conference on Machine Translation*, pages 355–413.

George Lakoff and Mark Johnson. 1980. *Metaphors we live by*, volume 1. University of Chicago press Chicago.

André Lefevere. 2016. *Translation, rewriting, and the manipulation of literary fame*. Routledge.

Chee Wee Leong, Beata Beigman Klebanov, Chris Hamill, Egon Stemle, Rutuja Ubale, and Xianyang Chen. 2020. A report on the 2020 vua and toefl metaphor detection shared task. In *Proceedings of the second workshop on figurative language processing*, pages 18–29.

Chee Wee Leong, Beata Beigman Klebanov, and Ekaterina Shutova. 2018. A report on the 2018 vua metaphor detection shared task. In *Proceedings of the workshop on figurative language processing*, pages 56–66.

Zhengjian Li and Lang Chen. 2025. Mind vs. machine: Comparative analysis of metaphor-related word translation by human and ai systems. *Training, Language and Culture*, 9(1):10–27.

Arle Lommel, Serge Gladkoff, Alan Melby, Sue Ellen Wright, Ingemar Strandvik, Katerina Gasova, Angelika Vaasa, Andy Benzo, Romina Marazzato Sparano, Monica Foresi, Johani Innis, Lifeng Han, and Goran Nenadic. 2024. The multi-range theory of translation quality measurement: MQM scoring models and statistical quality control. In *Proceedings of the 16th Conference of the Association for Machine Translation in the Americas (Volume 2: Presentations)*, pages 75–94, Chicago, USA. Association for Machine Translation in the Americas.

Arle Lommel, Hans Uszkoreit, and Aljoscha Burchardt. 2014. Multidimensional quality metrics (mqm): A framework for declaring and describing translation quality metrics. *Tradumàtica*, (12):0455–463.

Xiaofei Lu and Ben Pin-Yun Wang. 2017. Towards a metaphor-annotated corpus of mandarin chinese. *Language Resources and Evaluation*, 51(3):663–694.

Zhongtao Miao, Qiyu Wu, Kaiyan Zhao, Zilong Wu, and Yoshimasa Tsuruoka. 2024. Enhancing cross-lingual sentence embedding for low-resource languages with word alignment. In *Findings of the Association for Computational Linguistics: NAACL 2024*, pages 3225–3236.

Katarina Milenković, Miloš Tasić, and Dušan Stamenković. 2024. Influence of translation on perceived metaphor features: quality, aptness, metaphoricity, and familiarity. *Linguistics Vanguard*, 10(1):285–296.

Katarzyna Molek-Kozakowska. 2014. Coercive metaphors in news headlines: A cognitive-pragmatic approach. *Brno studies in English*, 40(1):149–173.

Peter Newmark. 1988. *A Textbook of Translation*. Prentice Hall, London.

Filippo Pallucchini, Lorenzo Malandri, Fabio Mercorio, and Mario Mezzanzanica. 2025. Lost in alignment: A survey on cross-lingual alignment methods for contextualized representation. *ACM Computing Surveys*, 58(5):1–34.

Jan Pedersen. 2017. How metaphors are rendered in subtitles. *Target*, 29(3):416–439.

Christina Schäffner. 2004. Metaphor and translation: some implications of a cognitive approach. *Journal of pragmatics*, 36(7):1253–1269.

Elena Semino. 2008. *Metaphor in discourse*. Cambridge University Press Cambridge.

Özgür Şen Bartan, Elif Arıca Akkök, and Kadir Yiğit Us. 2026. A comparative study of humans and machine learning in metaphor detection: Translations of legal metaphors in english and turkish hudoc judgments. *International Journal for the Semiotics of Law-Revue internationale de Sémiotique juridique*, pages 1–22.

Chung-ling Shih. 2012. A corpus-aided study of shifts in english-to-chinese translation of prepositions. *International Journal of English Linguistics*, 2(6):50.

Mark Shuttleworth. 2017. *Studying scientific metaphor in translation*. Routledge.

Marthe Smedinga, Alan Cienki, and Henk W de Regt. 2023. Metaphors as tools for understanding in science communication among experts and to the public. *Metaphor and the Social World*, 13(2):248–268.

Maria Carmen Staiano, Lifeng Han, Johanna Monti, and Francesca Chiusaroli. 2025. Italert: Assessing the quality of llms and nmt in translating italian emergency response text. In *Proceedings of Machine Translation Summit XX: Volume 1*, pages 566–577.

Gerard J Steen, Aletta G Dorst, J Berenike Herrmann, Anna A Kaal, and Tina Krennmayr. 2010a. Vu amsterdam metaphor corpus.

Gerard J Steen, Aletta G Dorst, Tina Krennmayr, Anna A Kaal, and J Berenike Herrmann. 2010b. A method for linguistic metaphor identification.

Gideon Toury. 2012. Descriptive translation studies–and beyond. *Benjamins Translation Library*.

Dita Trčková. 2011. Multi-functionality of metaphor in newspaper discourse. *Brno studies in English*, 37(1):139–151.

Raymond van den Broeck. 1981. The limits of translatability exemplified by metaphor translation. *Poetics Today*, 2(4):73–87.

Shun Wang, Ge Zhang, Han Wu, Tyler Loakman, Wenhao Huang, and Chenghua Lin. 2024. Mmte: Corpus and metrics for evaluating machine translation quality of metaphorical language. In *Proceedings of the 2024 conference on empirical methods in natural language processing*, pages 11343–11358.

Sum Wong and Qiliang Xu. 2025. Mapping metaphor research in translation and interpreting studies: a bibliometric analysis from 1964 to 2023. *Poznan Studies in Contemporary Linguistics*, 61(4):597–621.

Alicja Zajdel. 2022. Catching the meaning of words: Can google translate convey metaphor? In *Using Technologies for Creative-Text Translation*, pages 116–138. Routledge.

# A Contribution Highlights

The highlighted contribution (revised/rephrased based on the reviews we received) of this work is:

- While general-purpose MQM or BLEU/COMET scores measure overall sentence fluency and adequacy, they obscure fine-grained failures in figurative, non-literal translation. Adapting HOPE to focus specifically on metaphor severity *fills an important evaluation gap.*

- Translating full segments/documents before extracting target sentences for sentence-level annotation prevents models from being artificially constrained by isolated sentence boundaries.

- Post-editing system outputs to create parallel gold references from monolingual VUAMC/PSUCMC corpora provide a valuable future asset for English–Chinese MT research (resource available on our project GitHub page).

- We (Section 4.2.3) provide nuanced qualitative examples illustrating trade-offs between "rigid fact-tracking" (GoogleMT/GPT-5.4) and "over-flexible localization/hallucination" (Hunyuan-7B).

# B Rationale for News Corpus

The rationale to use the news domain is that news is rich in metaphors and it is not a widely studied domain for this topic yet. Corpus research has shown that, among academic articles, fiction, conversation, and news discourse, news texts contain the second-highest frequency of metaphor-related expressions, exceeding both fiction and conversational discourse (Steen et al., 2010a,b), which suggests that metaphor constitutes an important linguistic feature of news discourse. News translation features: compared with other discourse types, such as literary translation that often emphasizes stylistic and aesthetic effects (Bassnett, 2013), news translation primarily focuses on the rapid and effective communication of information to target audiences. To achieve this, news texts are often adapted to suit the communicative needs and reading conventions of target readers (Bielsa and Bassnett, 2008).

At the same time, news translation is not a fully objective transfer of information, but a process shaped by cultural, institutional, and ideological factors through selection and rewriting (Lefevere, 2016). As a result, information may be reorganized or reframed to align with the sociocultural and sociopolitical expectations of target audiences (Bielsa and Bassnett, 2008).

Within this context, metaphors in news discourse may help make complex events and issues more understandable and vivid to readers, while also carrying rhetorical, emotional, humorous, ironic, and ideological functions in the representation and framing of news events (Semino, 2008). Studies on media discourse further suggest that metaphors in news reporting may dramatize events, influence readers' responses, and guide interpretations of political and social issues (Trčková, 2011; Molek-Kozakowska, 2014). Metaphors in the news are therefore not merely decorative expressions, but important discourse tools that influence how events are presented and understood.

Overall, metaphor translation in news discourse warrants attention, since translation shifts or errors may alter the meanings and functions of source metaphors. Investigating translation errors and translation quality in news metaphor translation is therefore important for examining how metaphorical meanings and discourse functions are conveyed across languages in news communication.

# C Examples of Source Corpus and Related Work

## C.1 Metaphor Study Corpora

There are parallel corpora for metaphor studies, such as the aforementioned work from Wang et al. (2024) on English-Italian and English-Chinese (one-directional). This work adopts a heterogeneous multi-domain corpus for benchmarking metaphor-sensitive MT evaluation.

Regarding the monolingual corpus, VUAMC is the largest available English corpus word-level-annotated for all metaphorical language use, based on a systematic and explicit metaphor identification procedure (MIPVU) (Steen et al., 2010b).[3] It covers about 190,000 lexical units from a subset of four broad registers from academic texts, conversation, fiction, and news texts. The corpus was used for metaphor identification shared tasks (Leong et al., 2018, 2020), and the Fleiss' Kappa is over 0.8, which indicates good inter-annotator agreement (Steen et al., 2010a).

TroFi is one of the early datasets for distinguishing literal and nonliteral usages of verbs, constructed from the Wall Street Journal (news discourse) (Birke and Sarkar, 2006, 2007).[4] It consists of 3,727 English sentences covering 50 verbs from news texts. The metaphoricity of verb usage is annotated at the word level with a binary classification of literal vs nonliteral. Inter-annotator agreement is reported on a subset of 200 examples, with a Cohen's Kappa of 0.77, indicating relatively good agreement.

PSUCMC is a word-level-annotated Chinese dataset based on an adapted version of the Metaphor Identification Procedure Vrije Universiteit (MIPVU), whose reliability for Mandarin Chinese has been validated through inter-annotator agreement.[5] It consists of 30,012 words and covers three registers: academic discourse, fiction, and news. Fleiss' Kappa on this corpus is over 0.8, which indicates good inter-annotator agreement (Lu and Wang, 2017). For our work, we select the VUAMC and PSUCMC corpora for English and Chinese source language, respectively, because they used the same annotation strategy. Both corpora are annotated at the lexical-unit level following the Metaphor Identification Procedure Vrije Universiteit procedure (MIPVU)(Steen et al., 2010b) , which identifies metaphorical expressions through dictionary-based comparison between contextual and more basic meanings, further improvement/modification from metaphor identification procedure (MIP) proposed by Group (2007). They have similar discourse, and both are in the news domain. These features make them comparable to studying two languages. For Meta-HOPE, we use these two source corpora to carry out MT and post-editing to **generate a reference translation**, rather than relying on existing human reference translations. This is because metaphor translation often allows multiple valid and creative solutions, and a single reference cannot adequately represent all acceptable interpretations. Therefore, the focus is placed on analyzing system outputs rather than comparing them against a fixed gold standard. We will present our methodology in detail in the next section.

[3] VUAMC: `http://www.vismet.org/metcor/documentation/home.html`

[4] Trofi: `https://natlang.cs.sfu.ca/software/trofi.html`

[5] PSUCMC: `https://sites.psu.edu/xxl13/cmc/`

# D MetaHOPE Error Types with Examples (Phase-1)

We list MetaHOPE error types with examples from the pilot-study Phase-1 in Table 8.

# E Data Preprocessing Details

## E.1 Data Extraction and Filtering via POS

Only metaphorically used nouns, verbs, adjectives, and adverbs are included in the analysis. Grammatical function words, particularly prepositions, are excluded, as previous studies on English-Chinese translation have shown that such items frequently undergo omission or transformation due to structural differences between the two languages (Shih, 2012). These translation shifts are often caused by grammatical differences between English and Chinese rather than metaphor processing itself. Therefore, the analysis focuses on content words in order to better examine metaphor translation patterns. Chinese-specific Part-of-Speech (POS) categories in the original corpus annotation, such as vn (verb-noun) and i (idiom), were further normalized into the broader categories used in this study (noun, verb, adjective and adverb). Since category boundaries in Chinese are often flexible, the classification was determined based on the contextual syntactic function of each metaphorical expression within the sentence. Similarly, English-specific annotation categories in the original corpus, such as "N+N" and "V+AV", were also mapped onto these broader categories according to the contextual meaning and syntactic function of the metaphorical expression in context.

## E.2 Segment Processing

Context is particularly important for resolving ambiguity, maintaining coherence, and interpreting context-dependent expressions, especially for metaphors. Also, human translators do not usually translate sentences completely in isolation but rely on the surrounding context when producing translations (Gladkoff et al., 2026). Therefore, instead of getting translations of isolated sentences, the present study *provides larger text segments to translation models in order to better approximate real translation conditions.* To remain compatible with the HOPE-based annotation framework, the generated translations are subsequently segmented back into sentences and aligned at the sentence/segment level for later annotation and post-editing analysis. However, annotators are required to first read the full source text in order to build contextual understanding before conducting sentence/segment-level annotation and evaluation. For the segment selection, the study focuses on authentic news reports and excludes genres such as editorials and opinion pieces. Compared with opinion-oriented news discourse, hard news reporting generally prioritizes clarity and information delivery, allowing a more controlled comparison of metaphor translation across languages and systems. For both translation directions, segments were sampled from the corpora. Since not every sentence within a sampled segment contains metaphorical expressions, only sentences containing metaphorically used nouns, verbs, adjectives, or adverbs were included in the final analysis. A single sentence may also contain multiple metaphorical expressions.

The final dataset consists of 200 metaphor-containing sentences for each translation direction, to ensure that the analysis remained manageable while still allowing for reasonably reliable observations of translation quality patterns. Research on translation quality evaluation has found, through comparisons of different sample sizes, that a sample size of fewer than 200 sentences cannot statistically reflect the MT system quality in the translation quality evaluation task (Gladkoff et al., 2022). [6] Since the MetaHOPE design involves detailed manual metaphor annotation and qualitative error analysis, 200 sentences were considered an appropriate balance between analytical reliability and the practical feasibility of in-depth analysis.

Table 10 shows the statistics of the two extracted corpora we used for the pilot phase, including language type, segments, sentence length, metaphor-containing sentences, total MRWs, and amounts from each POS. The statistics on the deployment set (200 segments) are shown in Table 9.

## E.3 Aligning Metaphor Related Words (Source, MT.output)

In this section, we discuss the detailed framework for the Alignment of Metaphor-Related Words (MRWs) in Translation and its application to our task of preparing the corpus for MetaHOPE annotators.

The objective of the word-level alignment task is to identify how the metaphor-related meaning associated with an MRW is realized in translation. The task focuses on semantic-functional correspondence rather than strict lexical equivalence. Since metaphor-related meaning is often context-sensitive, semantically rich, and culturally dependent, metaphor translation frequently involves reformulation across linguistic and cultural systems. Therefore, it requires annotation to examine the target text to identify whether there is a target word, phrase, or larger textual unit that realizes the metaphor-related meaning associated with the source MRW. The alignment framework is informed by translation shift and equivalence literature in Translation Studies, which suggests that translation frequently involves structural, semantic, and pragmatic reformulation rather than strict formal correspondence (e.g., Catford, 1965; Baker, 1992; Chesterman, 1997). Particularly, paraphrasing, implicitation, restructuring, semantic reformulation, and pragmatic adaptation are common translational strategies that may affect how metaphor-related meaning is realized in the target text. As a result, metaphor-related meaning is not always realized through direct lexical correspondence in translation. Instead, it may be reformulated, redistributed across multiple units, implicitly conveyed through syntactic structure or discourse context, partially preserved, or omitted altogether. Alignment decisions in the present study are therefore made according to how

[6] Although this pilot study is based on a relatively small sample, it already reveals a diverse range of metaphor translation errors across all evaluated systems. These observations demonstrate the practical utility of MetaHOPE for systematically identifying, categorising, and analysing metaphor translation errors, while larger-scale experiments are left for future work.

| POS | token | words_label | words_gloss | type | subtype | sentence |
|---|---|---|---|---|---|---|
| v | 展开 | mrw | 开展，大规模地进行 | indir | conv | 这里不是商场，但竞争却已展开。 |
| v | 如 | mflag | | | | 聊着聊着，已近中午，两个厂家的咨询台前仍然人如潮涌。 |
| n | 方 | mrw | 交易、战斗或辩论的一个组成部分 | indir | conv | 一方，陕西＂长岭阿里斯顿＂。 |
| v | 吸引 | mrw | 将物体、力量或他人的注意力引到自己 | indir | conv | 无独有偶，雪花电器公司的２５１升大容积冰箱，也吸引了不少羡慕的目光。 |
| n | 方 | mrw | 交易、战斗或辩论的一个组成部分 | indir | conv | 两个厂家的产品一方典雅，一方豪华，姿质独具，在人头攒动中风头出尽。 |
| n | 方 | mrw | 交易、战斗或辩论的一个组成部分 | indir | conv | 两个厂家的产品一方典雅，一方豪华，姿质独具，在人头攒动中风头出尽。 |
| v | 出 | mrw | 出现;显露 | indir | conv | 两个厂家的产品一方典雅，一方豪华，姿质独具，在人头攒动中风头出尽。 |
| v | 获 | mrw | 得到;取得 | indir | conv | 前几天在东安市场展销评比中双获金奖，这次又在西单街头摆开擂台。 |
| v | 摆开 | mrw | | indir | widlii | 前几天在东安市场展销评比中双获金奖，这次又在西单街头摆开擂台。 |
| n | 擂台 | mrw | 旧时武术家比武艺的台子 | indir | cont | 前几天在东安市场展销评比中双获金奖，这次又在西单街头摆开擂台。 |

Figure 6: PSUCMC formatting.

| wordcat | lemma | word | metaphor | mettext | DELMET | sen |
|---|---|---|---|---|---|---|
| V | reveal | reveals | mrw | reveals | non-deliberate | Latest corporate unbundler reveals laid-back approach: Roland Franklin, who is leading a 697m pound break-up bid for DRG, talks to Frank Kane |
| N | casino | casino | mrw | casino | potentially deliberate | He rejects charges that he was partly responsible for the "casino atmosphere" that gripped US corporate life in the early1980s. |
| N | approach | approach | mrw | approach | non-deliberate | Latest corporate unbundler reveals laid-back approach: Roland Franklin, who is leading a 697m pound break-up bid for DRG, talks to Frank Kane |
| V | lead | leading | mrw | leading | non-deliberate | Latest corporate unbundler reveals laid-back approach: Roland Franklin, who is leading a 697m pound break-up bid for DRG, talks to Frank Kane |
| V | make | made | mrw | made | non-deliberate | IT SEEMS that Roland Franklin, the latest unbundler to appear in the UK, has made a fatal error in the preparation of his £ 697 m break-up bid for stationery and packaging group DRG. |
| AJ | fatal | fatal | mrw | fatal | non-deliberate | IT SEEMS that Roland Franklin, the latest unbundler to appear in the UK, has made a fatal error in the preparation of his £ 697 m break-up bid for stationery and packaging group DRG. |

Figure 7: VUAMC formatting.

metaphor-related meaning is represented in translation rather than through strict lexical matching.

Manual alignment: Pallucchini et al. (2025) argue that multilingual models still struggle with polysemy, homonymy, and language-specific semantic structures, while current alignment methods often rely on only "implicit and somewhat weak" correspondence signals between languages. Similarly, Miao et al. (2024) note that "the acquisition of token-level or word-level supervisory signals remains a challenging topic of ongoing discussion," indicating that reliable token-level semantic alignment is still unresolved in multilingual NLP systems. Therefore, even if auto-alignment tools are involved, manual verification and correction are still required. Based on the consideration above, this study adopts manual annotation directly to ensure context-sensitive and semantically informed alignment decisions.

**Alignment Principle with Examples** — In metaphor translation, metaphor-related meanings are realised through different translational patterns, as shown in Figure 8. MetaHOPE annotators need to examine the target text to determine how metaphor-related meaning is conveyed in translation using the prepared highlighted/aligned data from this guideline framework.

# F Detailed MetaHOPE Annotation Agreement - pilot v1

In Table 11, we list the "Per-error-type inter-annotator agreement" across 65 metaphor segments from the English-to-Chinese corpus from phase-1 of our pilot study.

# G Discussion

When a metaphor is translated as a paraphrase in the target language rather than as a metaphorical expression, whether it is subject to a translation-error penalty is a question. In our Pilot-v1 study on MetaHOPE annotation design, we included it as a penalty; however, it is not strictly applied by all annotators from our observation in the pilot-v1 study period. Adopted from literature discussion, e.g., (Toury, 2012), for updated annotation guidelines as described in Section J.2, we do not assign a penalty for this.

# H Prompts Used

For Hunyuan-MT-7B, the default prompt provided in the Hugging Face example is adopted to ensure standardized usage. As systems such as Google Translate and Hunyuan-MT-7B operate without external contextual information, the same prompt format is applied to GPT to control for prompt-related variation and ensure comparability across systems. The prompt example is as below:

"***Translate the following segment into Chinese, without additional explanation. [input texts]***"

# I Implications and Future Work

More broadly, MetaHOPE contributes toward bridging cognitive metaphor theory and empirical MT evaluation. Existing metaphor translation research has often emphasized translation strategies, metaphor preservation, or conceptual mapping, while MT evaluation research typically relies on broad sentence-level quality metrics. MetaHOPE offers a middle ground by enabling fine-grained, metaphor-sensitive error analysis that can systematically identify how metaphorical

Table 8: MetaHOPE error types with illustrative examples and explanations (pilot-phase1).

| Type | English Source | Problematic Translation | Suggested Translation | Explanation |
|---|---|---|---|---|
| **IMP** | *The government cracked down on protesters after the unrest.* | 抗议者在骚乱后被镇压。 | 政府在骚乱后镇压抗议者。 | The translation shifts the clause from active to passive voice, suppressing the agent (the government) and changing the distribution of agency, thereby altering communicative impact. |
| **RAM** | *The company is a black sheep in the field.* | 这个公司是行业里的问题公司。 | 这个公司是行业里的害群之马。 | The metaphor is paraphrased rather than adapted into an idiomatic Chinese equivalent. The figurative meaning is partially retained, but metaphorical and cultural adaptation is missing. |
| **MIS** | *The company is on its last legs after years of losses.* | 该公司在多年亏损后正处于它的最后几条腿上。 | 该公司在多年亏损后已处于倒闭边缘。 | The metaphor is translated literally into an unintelligible expression, causing loss of the intended meaning (near collapse or bankruptcy). |
| **STL** | *Officials lashed out at the decision, calling it irresponsible.* | 官员对这一决定表示不满，称其不负责任。 | 官员强烈抨击这一决定，称其不负责任。 | The translation weakens the metaphorical intensity and emotional force of the source expression, resulting in stylistic attenuation. |
| **PRF** | *The plan has gained momentum in recent weeks.* | 该计划在最近几周获得了动量。 | 该计划在最近几周获得了势头。 | The translation is understandable but unnatural in Chinese. A more idiomatic lexical choice would improve fluency and naturalness. |

Table 9: Corpus statistics of the metaphor translation dataset, including average sentence length and the distribution of metaphor-related word categories— for the MetaHOPE project deployment phase.

| **Statistic** | **VUAMC** | **PSUCMC** |
|---|---|---|
| Language | English | Chinese |
| Segments | 7 | 26 |
| Avg. Sentence Length (tokens) | 12.76 | 27.76 |
| Metaphor-containing Sentences | 200 | 200 |
| Total MRWs | 565 | 368 |
| Noun | 185 | 98 |
| Verb | 264 | 228 |
| Adj. & Adv. | 116 | 42 |

meaning is preserved, distorted, weakened, adapted, or lost during translation.

The present work represents a proof-of-concept study and opens several directions for future research. First, we plan to extend MetaHOPE to the full-scale test set and further refine annotation guidelines to improve inter-annotator agreement. Second, although this study focuses on the news domain, future work will examine the generalizability of MetaHOPE across other metaphor-rich domains, such as literary texts, fiction, academic discourse, political speeches, and science communication, potentially using corpora such as literary metaphor datasets and scientific communication corpora. Third, future analyses may investigate how different metaphor types—such as conventional vs. novel metaphors, deliberate vs. non-deliberate metaphors, or culture-specific vs. potentially universal metaphors—affect MT and LLM translation behavior. Finally, future work may explore semi-automatic or LLM-assisted metaphor annotation and alignment support, reducing human annotation costs while maintaining interpretability and reliability in metaphor-sensitive translation evaluation.

# J Dev Set Milestone-II: 2nd round annotations

## J.1 Improved Annotation Guidelines - V2/Phase2

Following the first pilot study on the Dev set, we further revised the annotator guidelines, whose details can be found at our open-source project website (https://github.com/Jiahui84/MetaHOPE). It includes:

- 1. the Annotation Task description.
- 2. Annotation Procedure (updated) with five steps.
- 3. Decision Rules (updated), attached below in Figure 9.
- 4. Error Categories (updated).
- 5. Severity Scale (updated).
- 6. Pilot Annotation.

Table 10: Statistics of the metaphor translation evaluation on the pilot dataset.

| **Statistic** | **VUAMC** | **PSUCMC** |
|---|---|---|
| Language | English | Chinese |
| Segments | 6 | 7 |
| Avg. Sentence Length (tokens) | 20.65 | 26.35 |
| Metaphor-containing Sentences | 20 | 20 |
| Total MRWs | 65 | 32 |
| Noun | 16 | 6 |
| Verb | 41 | 23 |
| Adj. & Adv. | 8 | 3 |

Table 11: Per-error-type inter-annotator agreement across 65 metaphor segments (pilot/dev-phase1).

| **System** | **Error** | $\alpha$ | **Weighted** $\kappa$ | **Pearson** $r$ | **Exact Agree.** |
|---|---|---|---|---|---|
| GoogleMT | IMP | 0.000 | 0.000 | NA | 98.5% |
| GoogleMT | RAM | -0.006 | 0.000 | NA | 96.9% |
| GoogleMT | MIS | 0.484 | 0.481 | 0.497 | 92.3% |
| GoogleMT | STL | -0.025 | 0.000 | NA | 90.8% |
| GoogleMT | PRF | 0.122 | 0.164 | 0.393 | 84.6% |
| GPT-5.4 | IMP | -0.008 | -0.016 | -0.016 | 96.9% |
| GPT-5.4 | RAM | -0.031 | -0.023 | -0.031 | 92.3% |
| GPT-5.4 | MIS | 0.424 | 0.421 | 0.458 | 92.3% |
| GPT-5.4 | STL | -0.034 | 0.000 | NA | 90.8% |
| GPT-5.4 | PRF | 0.039 | 0.113 | 0.305 | 76.9% |
| Hunyuan-LLM-7B | IMP | -0.017 | -0.014 | -0.024 | 93.8% |
| Hunyuan-LLM-7B | RAM | -0.007 | 0.000 | NA | 96.9% |
| Hunyuan-LLM-7B | MIS | 0.524 | 0.535 | 0.646 | 83.1% |
| Hunyuan-LLM-7B | STL | -0.023 | 0.000 | NA | 92.3% |
| Hunyuan-LLM-7B | PRF | -0.091 | -0.010 | -0.057 | 76.9% |

- 7. Annotation Requirements.

As shown in Figure 9, the annotation procedure includes: Step 1: Read the Full Context (texts are in the folder and are named with docid). Read the entire article to understand the overall topic, discourse context, communicative purpose, and intended meaning of the metaphor.

Step 2: Determine Whether an Error Exists. Question: Does the translation contain a metaphor-related translation error? Options: No Error / Error. If no error is identified, no further annotation is required.

Step 3: Identify Error Type(s). If an error exists, select the applicable error category or categories from the framework below. If multiple independent errors are present, multiple categories may be assigned. For each annotated instance, it can be one error type, or multiple error types, e.g. MIS and IMP. Assign the corresponding error severity levels for all possible error types, wherever needed.

Step 4: Assign Severity. Assign a severity score to each error type that you selected. (2,4,6,8,10): using the severity scale.

Step 5: Produce a human gold standard reference; can use post-editing or totally from your own. If an error is identified, annotators may provide a suggested revision in the Human Gold Standard column. The revision may be either selected from one of the model translations or produced by post-editing one of the model outputs.

## J.2 Error Type Definitions and Examples (updated)

We updated the error type definition from the perspective of metaphor-based translation studies. We give bidirectional translation examples en<=>zh, with suggested references and explanations.

### J.2.1 IMP (Impact)

An Impact Error occurs when the translation conveys the basic meaning of a metaphor but fails to preserve its intended communicative impact, such as its persuasive force, emotional effect, evaluative stance, or emphasis, thereby producing a substantially weaker or different effect on the target audience.

Example 1 (EN-ZH): EN: Officials lashed out at the decision. ZH: 官员对这一决定表示不满 (gloss: Officials expressed dissatisfaction at the decision.) ->suggest translation: 官员强烈抨击这一决定 (gloss: strongly at-

1) Direct lexical correspondence
Example:
EN: … the more luxurious the luncheon rooms at **headquarters,** the more inefficient the business".
ZH: 总部餐厅越豪华，企业的运营效率就越低。"

ZH: …先后关闭了它在台湾和南朝鲜的分厂，而将资金转移到中国。
EN: … **closed** its factories in Taiwan and South Korea and transferred its funds to China.

2) Paraphrastic realization
Example:
EN: … could not expect everybody to "**goose-step**" in the same direction …
ZH: 也不能指望所有党员都会盲目追随党的领导方向。

ZH: 于是刘庄子人打定主意：请不着就〞抢〞。
EN: Therefore, the villagers decided: if they couldn't invite him, they would "**seize the opportunity**" themselves.

3) Distributed realization across multiple units
Example:
EN: … financial considerations would not **stand** in the way of implementing…
ZH: …财政因素不会成为实施安东尼·希登（Anthony Hidden）提出的93项整改建议的障碍。

ZH: …不少农家陷入断炊困境
EN: … **putting** them **in** a dire situation.

4) Implicit realization through syntactic structure (e.g., context needed)
Example:
EN: Nor, it seems, **is** anyone else.
ZH: 但似乎没有其他企业愿意承担全国范围内的推广任务。

5) Partial metaphorical reformulation
Example:
EN: Britain still cannot decide when to play the mandarinate **game** of silence …
ZH: 英国至今仍无法决定何时应该对中国保持沉默……

ZH: …冰箱出了问题，打个电话，随叫随到......
EN: … can call for immediate assistance if their refrigerators **need repair**.

6) Omission
Example:
EN: All of which is not just a problem for a properly **penitent** British Rail: it's a problem for the government too.
ZH: 这不仅是英国铁路的问题，更是政府的问题

ZH: … 冰箱冷冻室容积达96升
EN: … its 96-liter freezer capacity

Figure 8: Metaphor translation alignment examples realized through different translational patterns

tack) Explanation: The translation preserves the basic meaning but weakens the metaphorical force. While lashed out conveys a forceful verbal attack, 表示不满 merely expresses dissatisfaction, reducing the persuasive force and rhetorical impact. The meaning is retained, but the communicative impact of the metaphor is weakened, resulting in an Impact Error.

Example 2 (ZH-EN): ZH: 该公司裁减了上千个岗位。(gloss: The company cut thousands of jobs.) EN: Thousands of jobs were cut. ->suggest translation: The company slashed thousands of jobs. Explanation: The translation preserves the basic meaning but shifts the communicative emphasis. By changing the focus from the company as the responsible actor to the affected jobs, it foregrounds the consequence rather than the action. The intended rhetorical emphasis is therefore altered, resulting in an Impact Error. *Note: A change in voice (e.g., active vs. passive) should not be treated as an Impact Error unless it substantially alters the intended communicative emphasis or rhetorical effect of the metaphor.

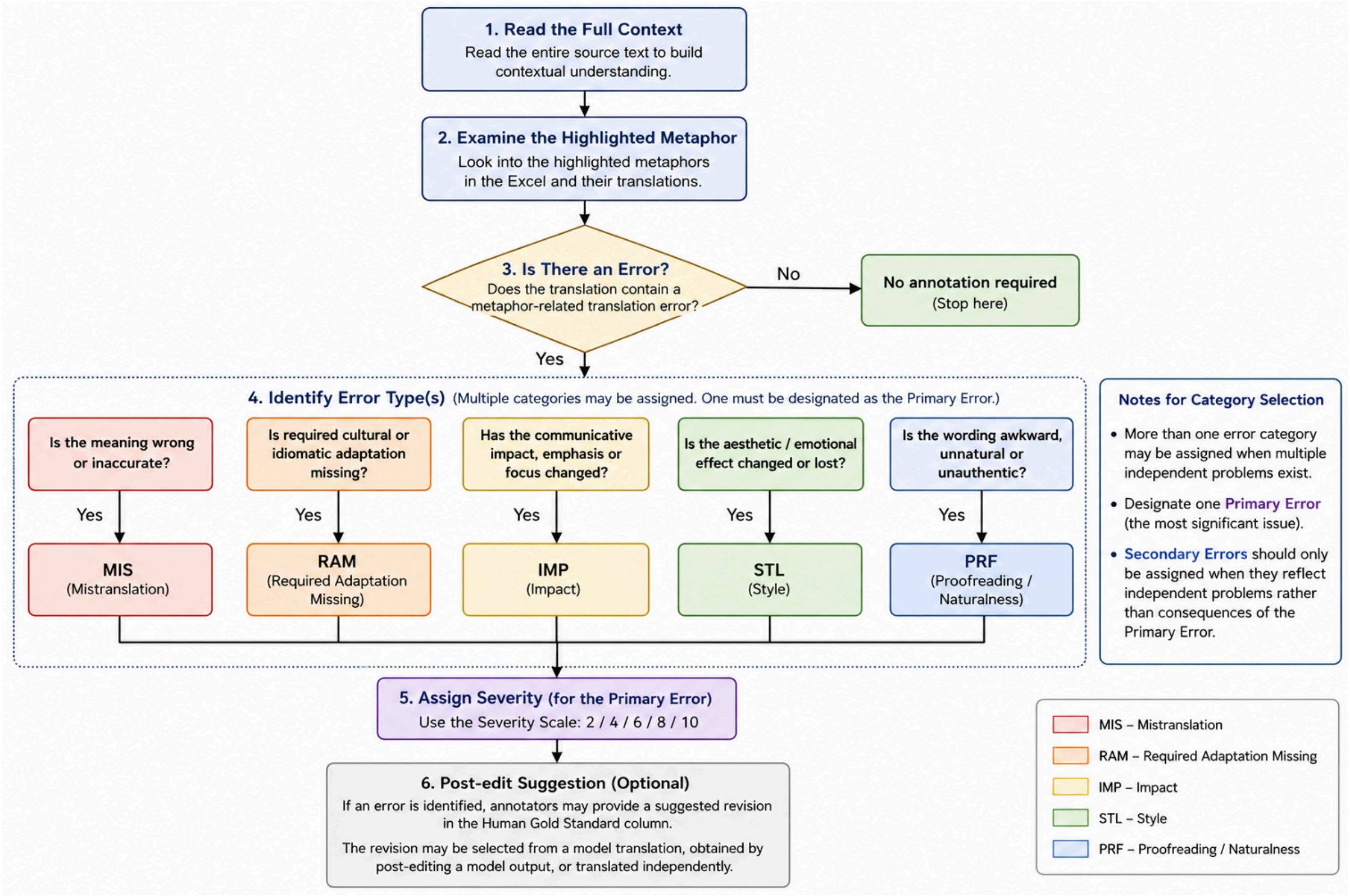

Figure 9: MetaHOPE Annotation Procedure via Decision Tree Illustration

### J.2.2 RAM (Required Adaptation Missing)

A Required Adaptation Missing (RAM) Error occurs when a source metaphor requires adaptation to target-language metaphorical, idiomatic, or cultural norms, but the necessary adaptation is not made.

EN: The plan has gained momentum in recent weeks. ZH: 该计划在最近几周获得了动量。(gloss: The plan has gained momentum in recent weeks) -> suggest translation: 该计划在最近几周势头上涨 (gloss: momentum) Explanation: The translation preserves the intended meaning but does not adopt the conventional metaphorical expression used in Chinese news discourse. Chinese news reports conventionally use 势头 (upward momentum or gaining strength) to describe increasing progress or development. By contrast, 获得了动量 (gained momentum) is a literal rendering of the English expression that is not idiomatic in Chinese. The required adaptation to target-language metaphorical convention is therefore missing, resulting in a Required Adaptation Missing Error.

ZH: 该政策给经济发展注入了一针强心剂。(gloss: The policy gave the economy a much-needed boost.) EN: The policy boosted the economy. -> suggest translation: The policy gave the economy a much-needed shot in the arm. Explanation: The translation preserves the intended meaning but omits the conventional English metaphor. English commonly uses a shot in the arm to express renewed strength or encouragement, whereas boosted the economy paraphrases the meaning without preserving the established metaphorical expression. The required target-language adaptation is therefore missing, resulting in a RAM Error.

### J.2.3 MIS (Mistranslation)

A Mistranslation Error occurs when the translation distorts the intended meaning of the source metaphor, resulting in an accuracy error and a different semantic interpretation in the target text. This typically occurs because the metaphorical interpretation, conceptual mapping, or contextual meaning is not correctly transferred.

EN: The company is on its last legs after years of losses. ZH: 该公司在多年亏损后正处于它的最后几条腿上。(literal translation: on its last legs) -> suggest translation: 该公司在多年亏损后正处于倒闭边缘 (gloss: The company is on the edge of bankruptcy after years of losses.) Explanation: The source metaphor maps physical weakness (last legs) onto impending failure. This conceptual mapping is not preserved in the literal Chinese translation, which instead prompts a literal interpretation and fails to convey the intended metaphorical meaning.

ZH: 他总喜欢给别人穿小鞋。(gloss: He always makes other people wear small shoes i.e., deliberately makes things difficult for them.) EN: He always makes people wear small shoes. -> suggest translation: He always makes life difficult for other people. Explanation: The Chinese idiom “穿小鞋 (wear small shoes)” is a conventional metaphor meaning to deliberately create difficulties for someone or treat them unfairly. The English translation interprets the expression literally as making people wear small shoes, rather than conveying its intended metaphorical meaning. As a result, the metaphorical interpretation is lost, producing a different semantic interpretation and resulting in a Mistranslation Error.

#### J.2.4 STL (Style)

A Style Error occurs when the translation preserves the metaphor' s basic referential meaning but alters the stylistic realization of the metaphor: register, tenor, voice, figurative vividness, or genre-appropriate rhetorical effect, thereby performing a different expressive or interpersonal function from the source.

Example: EN: The market has slipped into a casino atmosphere. ZH: 市场变得投机性越来越强。(gloss: The market has become increasingly speculative.) ->suggest translation: 市场充斥着赌场般的投机氛围。(gloss: The market is filled with a casino-like speculative atmosphere.) Explanation: The translation "投机性 (speculative)" preserves the intended meaning of "casino" but paraphrases the metaphor into a neutral expression, resulting in the loss of its figurative imagery and rhetorical style. Since metaphor serves both semantic and stylistic functions, such cases can be classified as Style Errors when the meaning is retained but the metaphorical expression is not.

ZH: 市场出现明显回暖。(gloss: The market has shown a clear warming trend.) EN: The market rose like a phoenix from the ashes. ->suggest translation: The market showed clear signs of recovery. Explanation: The translation "rose like a phoenix from the ashes" conveys part of the meaning of "回暖 (market) recovery" but adopts a highly literary and fictional metaphor that is stylistically inappropriate for the news genre. Although the propositional meaning is preserved, the target text shifts from an objective journalistic style to a poetic narrative voice, resulting in a Style Error.

#### J.2.5 PRF (Proofreading / Naturalness)

A Proofreading / Naturalness Error occurs when the metaphor is translated with basically correct meaning and appropriate metaphorical interpretation, but the target expression is awkward or unnatural. The error affects the fluency or acceptability of the translation rather than meaning transfer.

Example: EN: Inflation is eating into household incomes. ZH: 通货膨胀正在吃进家庭收入。(gloss: Inflation is eating into household incomes.) -> suggest translation: 通货膨胀正在侵蚀家庭收入。(gloss: Inflation is eroding household income.) Explanation: The translation preserves both the intended meaning and the metaphorical interpretation of the source expression. However, the wording "吃进 (eat into)" is awkward and does not conform to natural Chinese usage. A professional translation would more naturally express this idea as "侵蚀 (erode)", improving fluency and acceptability without changing the underlying meaning or metaphorical interpretation. The problem therefore lies in target-language naturalness rather than meaning transfer, resulting in a Proofreading Error.

ZH: 她决定跳槽到一家更有活力的初创公司。(gloss: She decided to jump trough to a more dynamic startup ie, change jobs for a better opportunity.) EN: She decided to jump trough to a more dynamic startup. -> suggest translation: She decided to switch jobs to a more dynamic startup. Explanation: The translation "jump trough" conveys the intended meaning of "跳槽 (jump trough i.e. changing jobs for a better opportunity)" and preserves the underlying metaphorical interpretation. However, "jump trough" is an awkward and unnatural expression in English, although its intended meaning can be inferred from the literal wording. Although the propositional meaning is preserved, the target expression requires only linguistic refinement rather than a change in meaning, resulting in a Proofreading Error.

Table 12: Ordinal Krippendorff's $\alpha$ by system and MetaHOPE error type.

| System | IMP | RAM | MIS | STL | PRF |
|---|---|---|---|---|---|
| GoogleMT | .092 | .125 | .055 | −.019 | **.179** |
| GPT-5.4 | −.039 | **.173** | −.006 | −.026 | .034 |
| Hunyuan-7B | .010 | .019 | **.112** | −.046 | **.140** |

Negative / Low / Moderate / Higher agreement

### J.3 Severity Scale (updated)

0- no error (can leave the Excel corresponding cell empty as it is)

2 –Minor. A minor issue that has little impact on the overall translation. The metaphor remains largely successful.

4 –Moderate. A noticeable error that partially affects the translation quality or metaphor realization, but the overall message remains clear.

6 –Major. A substantial error that clearly weakens the translation. The metaphorical meaning, function, stylistic effect, or naturalness is significantly affected and requires revision.

8 –Severe. A serious error that causes major loss or distortion of the metaphor's intended meaning, function, stylistic effect, or naturalness.

10 –Critical. A critical error that fails in metaphor translation. The intended metaphorical meaning, function, stylistic effect, or naturalness in the target-language realization is largely lost or seriously misleading.

### J.4 More Detailed Outputs on Pilot-v2

We further draw the heat map of Per-system inter-annotator agreement by MetaHOPE error type in Table 12. The heat map makes the main patterns immediately visible: GoogleMT–PRF, GPT-5.4–RAM, and Hunyuan-7B–MIS/PRF show the strongest agreement, while STL is negative for all three systems. This suggests that: a) GoogleMT' s naturalness problems were comparatively recognisable under PRF. b) GPT-5.4' s missing adaptation cases were comparatively recognisable under RAM. c) Hunyuan' s meaning-related problems were most consistently captured under MIS, although agreement was still weak.

### J.5 Examples of major disagreement

#### J.5.1 whether the metaphor meaning was mistranslated

This is both a detection disagreement and a category-boundary disagreement. We display in Table 13 for

Table 13: Illustrative example of inter-annotator disagreement on a metaphor translation. The example demonstrates disagreement in (i) error detection, (ii) error categorisation, and (iii) severity assessment, motivating the refinement of the MetaHOPE annotation guidelines.

| **ID** | `news_01_v.4_10_17` |
|---|---|
| **System** | Hunyuan-7B |
| **Source excerpt (ZH)** | 为迎接竞赛做题三千多道 |
| **Literal gloss** | "(He) completed/solved more than 3,000 practice questions in preparation for the competition." |
| **System output (Hunyuan-7B)** | *"She prepared over three thousand practice problems for the competition."* |
| **Metaphor-related expression** | 做题 ("to do/solve practice questions") |

| **Annotator** | **IMP** | **RAM** | **MIS** | **STL** | **PRF** | **Total** | **Possible interpretation** |
|---|---|---|---|---|---|---|---|
| Annotator C | 2 | 0 | 0 | 0 | 0 | 2 | Minor impact shift; translation generally acceptable. |
| Annotator D | 0 | 0 | 0 | 0 | 0 | 0 | No translation error identified. |
| Annotator E | 0 | 0 | 8 | 4 | 0 | 12 | Severe mistranslation of the metaphor-related meaning together with a noticeable reduction in metaphorical force. |
| Annotator A | 0 | 0 | 8 | 0 | 2 | 10 | Severe mistranslation with minor target-language awkwardness. |
| Annotator F | 0 | 0 | 2 | 0 | 0 | 2 | Minor semantic shift; the overall meaning is largely preserved. |

**Discussion.** This example illustrates three major sources of disagreement observed in the pilot study. First, annotators differed on *error detection*: whether the translation *prepared* adequately conveys the meaning of the Chinese expression 做题 ("to do/solve practice questions"), with one annotator assigning no error while others identified varying degrees of error. Second, annotators differed on *error categorisation*: some regarded the problem as a change in communicative impact (IMP), whereas others classified it as a mistranslation (MIS) with or without additional stylistic loss (STL) or proofreading issues (PRF). Finally, even when annotators agreed that a mistranslation was present, they assigned substantially different *severity levels* (minor vs. severe). These observations motivated the refinement of the MetaHOPE annotation guidelines by providing clearer decision rules for distinguishing IMP from MIS and for calibrating severity scores.

a case study in this category, using the source sentence "为迎接竞赛做题三千多道" extracted from the full segment. In this example, GoogleMT transalted as "did", GPT gave "solved", while Hunyuan used "prepared". The translation of 做 (题) ("work through/-complete practice problems") generated substantial disagreement among annotators. Annotator D considered "did" (from GoogleMT) a severe RAM error (8), possibly suggesting it's better way in target language to express this. Annotator F regarded "solved" (from GPT) as a moderate RAM error (4), while classifying "prepared" (from Hunyuan) as a minor MIS error (2). Annotator C interpreted both "did" and "prepared" as minor IMP errors (2) and proposed "work through" as the gold standard translation. In contrast, Annotator A argued that "solved" constituted a minor MIS error (2), noting that doing practice problems does not necessarily imply solving them, and classified "prepared" as a severe MIS error (8) because it changes the event from working through problems to preparing problems. Annotator E similarly assigned "prepared" a severe MIS error (8), arguing that it misinterprets 做题 as preparing exercises, thereby reversing the agent role. Annotator E additionally assigned a Style error (4), commenting that the wording was overly formal, weakened the conversational tone of the original dialogue, and could be expressed more naturally.

### J.5.2 omission or legitimate paraphrase

This is related to our discussion in Section G. As an example, the Words: 方向 and 拉 from token IDs: news_03_v.4_74_25 and news_03_v.4_74_27 from the source sentence "坚持送戏下乡的正确方向", and the full segment: "这台 " 东风 " 加长大卡车，是自治区党委和人民政府为支持宁夏话剧团坚持送戏下乡的正确方向, 拉专款购置的。" System: Hunyuan-7B translated "...to support the troupe's efforts in bringing performances to rural areas." For each metaphor, four annotators assigned either 0 or 2, while Annotator A assigned MIS = 10. The key question is whether Hunyuan's sentence adequately conveys the policy-oriented expression about 坚持送戏下乡的**正确方向**. It preserves the practical activity of bringing performances to rural areas but omits or neutralises "the correct direction." This example exposes an unresolved rule: Is omission of political or ideological framing an Impact error? Is it Mistranslation because part of the propositional meaning disappears? Or is it an acceptable target-language reformulation?

Another example in this category is the word "抓紧" from the segment "昨天从沙特返回的 9 名内阁成员已着手工作，抓紧日常生活的恢复工作。"[7] GoogleMT:

[7]GPT5.4 had the correct translation: The nine cab-

"Nine cabinet members who returned from Saudi Arabia yesterday have begun work to restore daily life. [omission]" Hunyuan-7b: "Nine cabinet members who returned from Saudi Arabia yesterday have begun to work on restoring daily life. [omission]" Different error types and severity levels were given by annotators — Annotator C: IMP(4). Annotator A: IMP(4), Annotator F: MIS(2), Annotator E: IMP(2), Annotator D: MIS(8).

#### J.5.3 metaphor alignment when the meaning is distributed

As an example, the word "进入" from segment 金文秀和朴柱奉是在半决赛中以２：１（１６：１７，１５：６，１５：１１）击败印度尼西亚的洪忠中和罗天宁之后**进入决赛**的。System: Hunyuan-7B translated into "Kim Mun-soo and Park Ju-bong had previously defeated Indonesia' s Hong Zhongzhong and Luo Tianning with a score of 2:1 (16:17, 15:6, 15:11) in the semifinals." The annotators assigned RAM = 2, IMP = 4, MIS = 10, MIS = 2, or no error. The likely issue is that 进入决赛 is not translated through a direct equivalent of "enter" ; instead, qualification is *inferred* from "defeated ⋯in the semifinals." The broader sentence may convey the result, although the individual metaphor-related word has no direct target counterpart.

This reflects that the further refinement of annotation guideline can be covered by an explicit alignment rule: e.g., "A metaphor-related word does not require a direct lexical counterpart when its meaning is correctly realised through another phrase or through the sentence structure. Do not penalise distributed or implicit realisation unless relevant meaning is lost."

## K Indications from Pilot-Phase2: 2nd round annotations

The updated guidelines appear to provide the clearest common ground for **MIS** and **RAM**, but they have not yet solved three major problems: 1) Error detection threshold: Some annotators penalise a questionable rendering while others assign zero. 2) MIS versus IMP/STL/PRF: The same issue is frequently placed in different categories. 3) Severity calibration: Annotators disagree strongly between minor, major, severe, and critical, particularly for partial meaning loss. These empirical-based findings shed some light that it is indeed a challenge for metaphor-focused translation error annotations, which still sets a bottleneck in the field.

## L Metric Related Works

For further reading, the HOPE metric has been adapted to other domains and languages, such as the crisis translation domain in Italian (Staiano et al., 2025) and the Arabic language translation Ara-HOPE (Alabdullah et al., 2026).

inet members who returned from Saudi Arabia yesterday had already begun work, pressing ahead with efforts to restore normal daily life.